\documentclass[preprint,12pt]{elsarticle}

\usepackage{amsmath,amssymb,amsfonts}
\usepackage{natbib}
\setcitestyle{authoryear,round}
\usepackage{subfigure}
\usepackage{threeparttable}
\usepackage{graphicx}
\usepackage{textcomp}
\usepackage{xcolor}
\usepackage{cases}
\usepackage{booktabs} 
\usepackage{multirow}
\usepackage{hyperref}
\usepackage{svg}
\usepackage{bbm}
\usepackage{algorithm}
\usepackage{algorithmic}
\hypersetup{hypertex=true,
colorlinks=true,
linkcolor=blue,
anchorcolor=blue,
citecolor=blue}

\journal{Engineering Applications of Artificial Intelligence.}

\begin{document}

\begin{frontmatter}

\title{Robot Crowd Navigation in Dynamic Environment
with Offline Reinforcement Learning}

\author[label1,label2]{Shuai~Zhou}
\author[label1,label2,label3]{Hao~Fu\corref{cor1}}
\author[label1,label2]{Haodong~He}
\author[label1,label2]{Wei~Liu}
\address[label1]{School of Computer Science and Technology, Wuhan University of Science and Technology, Wuhan 430081, China}
\address[label2]{Hubei Province Key Laboratory of Intelligent Information Processing and Real-time Industrial System, Wuhan 430081, China}
\address[label3]{Hubei Key Laboratory of Digital Textile Equipment, Wuhan 430200, China}
\tnotetext[t1]{This work was supported in part by the National Natural Science Foundation of China under Grant 62303357, in part by the Natural Science Foundation of Hubei Province under Grant 2023AFB109, in part by the Knowledge Innovation Project of Wuhan under Grant 2022010801020315, and in part by the Hubei Provincial Advantaged Characteristic Disciplines (Groups) Project of Wuhan University of Science and Technology under grant 2023D031.}
\cortext[cor1]{Corresponding author: Hao Fu, Email: fuhao@wust.edu.cn.}

\begin{abstract}
Robot crowd navigation has been gaining increasing attention and popularity in various practical applications.
In existing research, deep reinforcement learning has been applied to robot crowd navigation by training policies in an online mode. However, this inevitably leads to unsafe exploration, and consequently causes low sampling efficiency during pedestrian-robot interaction. To this end, we propose an offline reinforcement learning based robot crowd navigation algorithm by utilizing pre-collected crowd navigation experience. Specifically, this algorithm integrates a spatial-temporal state into implicit Q-Learning to avoid querying out-of-distribution robot actions of the pre-collected experience, while capturing spatial-temporal features from the offline pedestrian-robot interactions. Experimental results demonstrate that the proposed algorithm outperforms the state-of-the-art methods by means of qualitative and quantitative analysis.
\end{abstract}

\end{frontmatter}

\section{Introduction}
Recently, the research on mobile robots has received everincreasing popularity from various practical applications, such as search and rescue robots, service robots, warehouse robots, etc.  As one of their hot research topics, robot navigation aims at autonomously planning the path from the starting position to the target while avoiding obstacles in the unknown environment. Some challenges arise when the robot navigates in the crowd environment, such as airports, hospitals, and shopping centers.

In existing research, deep reinforcement learning (DRL) \citep{mnih2013playing} has been applied to the task of robot crowd navigation. It provides a method for learning navigation policies from raw sensory (such as laser scans \citep{long2018towards} or images \citep{sathyamoorthy2020densecavoid}) or agent-level state representations \citep{chen2017decentralized, chen2017socially}. Specifically, in DRL methods, the navigating robot needs to interact frequently with pedestrians in an enriched crowd environment, collecting sufficient exploration samples through iterative trials to enhance its navigation capabilities. However, under this training mode, inadequate learning makes it challenging to acquire efficient learning experiences, leading to a low utilization rate of sampled data by the robot. This issue is particularly severe in the early stages of training. Furthermore, in order to effectively enhance the robot's navigation capabilities, robots typically discard outdated experiences during training. Due to these reasons, the sampling efficiency during pedestrian-robot interactions in DRL-based robot crowd navigation is extremely low. Additionally, insufficient learning inevitably results in the robot's unsafe exploration, posing a collision threat to both the robot and pedestrians.

Motivated by above observations, this paper proposes an offline reinforcement learning based robot crowd navigation algorithm to deal with unsafe exploration of the pedestrians and the robot, and low sampling efficiency during pedestrian-robot interaction. Furthermore, in order to obtain effective representations of pedestrian-robot interaction, a ST2-ORL algorithm is developed by integrating the spatial-temporal state transformer (ST2) \citep{yang2023st} into offline reinforcement learning. The main contributions of this paper are shown as follows:
\begin{enumerate}[1)]
\item Our proposed robot crowd navigation algorithm modifies the loss function in the SARSA-style objective to obviate querying out-of-distribution robot actions of the pre-collected crowd navigation experience.
\item Combining implicit Q-Learning and spatial-temporal state transformer allows the robot to learn effectively capturing spatial-temporal features from the offline pedestrian-robot interactions.
\item Qualitative and quantitative evaluation on different highly dynamic environments and comparison against a baseline DRL approach without semantic information
\end{enumerate}

The organization of this paper is as follows. Section \ref{2} introduces related works. Section \ref{3} first formulates the robot crowd navigation problem, then proposes an offline reinforcement learning for robot crowd navigation, and finally provides details about the ST2-ORL architecture. Then, the qualitative and quantitative results of experiments and simulation are presented in Section \ref{4}. Finally, Section \ref{5} concludes this paper.

\section{RELATED WORKS}\label{2}

\subsection{Reaction-based Methods}
The problem of robot navigation in dynamic environments, especially crowd navigation, has been extensively explored for several decades. Social Force \citep{ferrer2013robot, helbing1995social} is an algorithm that simulates the movement and behavior of individuals in the crowd environment based on the forces of interaction between pedestrians. Other works consider the velocity of obstacles, such as Velocity Obstacles (VO), which is necessary in dynamic environments. Reciprocal Velocity Obstacles (RVO) \citep{van2008reciprocal} and Optimal Reciprocal Collision Avoidance (ORCA) \citep{van2011reciprocal} model crowds as velocity obstacles and assume that agents avoid each other based on reciprocal rules. However, these reaction-based methods, due to their failure to consider the future evolution of neighboring entities, are short-sighted in terms of time and struggle to handle complex conditions, particularly when the crowd size expands.

\subsection{Trajectory-based Methods}
Trajectory-based methods first predict the expected trajectories of each pedestrian in a crowd and then plan a feasible path for the robot \citep{chen2021interactive, ferrer2014proactive}. Trajectory prediction allows the robot to plan navigation paths under longer time horizons, leading to better navigation outcomes. However, predicting the trajectories of individuals within a crowd and performing online path searching from a large state space entail substantial computational expenses. Additionally, the $freezing\ robot$ problem \citep{trautman2010unfreezing} in trajectory-based methods is also challenging to address.

\subsection{Learning-based Methods}
With the development of deep learning, in some works, imitation learning has been utilized to learn policies from demonstrated expected behaviors. In \citet {long2017deep}, the target behavior is learned through demonstration data. This method often performs poorly on suboptimal experiences and heavily relies on expert-collected datasets.

Another group of works utilizes deep reinforcement learning (DRL) methods, where deep neural networks are employed to map input states to robot actions. Based on the input states, DRL-based methods can be further categorized into end-to-end methods and agent-level methods. End-to-end methods take raw perceptual information, such as radar or images, as inputs to deep neural networks. Their challenge lies in the lack of explicit high-level information, particularly the dynamic interaction characteristics between the robot and pedestrians, which are crucial for modeling crowd interactions.

In contrast, agent-level DRL-based methods use agent-level state representations, such as the positions and velocities of the robot and pedestrians, as inputs to deep neural networks. Based on agent-level interaction information, they can learn explicit models of crowd interaction, such as social norms \citep{chen2017socially}, spatial-temporal graphs \citep{liu2021decentralized, wang2023navistar}, and attention weights \citep{chen2019crowd, liu2023intention, zhou2022robot}. Furthermore, benefiting from low-dimensional yet high-level state representations, agent-level methods can achieve real-time performance on low-cost robots and offer flexibility in choosing perceptual sensors.

However, due to the limitations of online training, it is inevitable that all these methods require frequent interaction with the environment to collect data for training the robot. Consequently, safety and cost issues arising from collisions between the navigating robot and pedestrians during the exploration process are unavoidable. Additionally, the low sampling efficiency during pedestrian-robot interaction is challenging to address.
Therefore, this paper focuses on this aspect and investigates the problem of robot crowd navigation using offline reinforcement learning driven by pre-collected datasets.

\section{METHODOLOGY}\label{3}
\subsection{Robot Crowd Navigation Modeling} \label{modeling}
The robot crowd navigation problem can be represented as a Markov Decision Process (MDP). The MDP is described as a tuple ($\mathcal{S}$, $\mathcal{A}$, $\mathcal{P}$, $\mathcal{R}$, $\gamma$), where $\mathcal{S}$ is the state space of the world, $\mathcal{A}$ is the action space of the robot, $\mathcal{P} : \mathcal{S} \times \mathcal{A} \rightarrow \mathcal{S}$ is the state transition probability, $\mathcal{R}$ is the reward function, and $\gamma \in(0, 1)$ is the discount factor. Building upon these fundamental concepts, the following provides a reinforcement learning formulation for the robot crowd navigation:

\textbf{State Space $\mathcal{S}$.} The environment of the task is described using a 2D coordinate system.
The state space at each time step consists of observable and unobservable states of the agents (robot and pedestrians). The observable part includes the position $\boldsymbol{\mathrm{p}}=[p_x, p_y] \in \mathbb{R}^2$, velocity $\boldsymbol{\mathrm{v}}=[v_x, v_y] \in \mathbb{R}^2$, collision radius $r \in \mathbb{R}$, and the unobservable part includes the goal position $\boldsymbol{\mathrm{p}}_g =[p_{gx}, p_{gy}] \in \mathbb{R}^2$, preferred velocity $v_{pre} \in \mathbb{R}$, and heading angle $\psi \in \mathbb{R}$.

This work follows \citet{chen2017decentralized} to parameterize the states of the robot and the pedestrians centered around the robot. Then, influence of the absolute position on decision-making is eliminated. Therefore, the global state $s_t$ can be represented as:
\begin{align}
    s_t^r &= [d_g, v_{pre}, \psi, r, v_x, v_y] \in \mathbb{R}^6,\\
    s_t^n &= [\widetilde{p}_x^n, \widetilde{p}_y^n, \widetilde{v}_x^n, \widetilde{v}_y^n, r^n, d^n, r+r^n]\in \mathbb{R}^7,\\
    s_t &= [s_t^0, s_t^1,..., s_t^n],
\end{align}
where $s_t^r$ and $s_t^n$ are the states of the robot and the $n$-th pedestrian at time $t$, $d_g=\Vert \boldsymbol{\mathrm{p}}_g -\boldsymbol{\mathrm{p}} \Vert_2$ is the robot’s
distance to the goal, $d^n=\Vert \boldsymbol{\mathrm{p}} -\boldsymbol{\mathrm{p}}^n \Vert_2$ is the robot’s distance to
the pedestrian $n$, and the relative position and relative velocity of the n-th pedestrian are $\boldsymbol{\mathrm{\widetilde{p}}}^n = [\widetilde{p}_x^n, \widetilde{p}_y^n]$ and $\boldsymbol{\mathrm{\widetilde{v}}}^n = [\widetilde{v}_x^n, \widetilde{v}_y^n]$, respectively.

\textbf{Action Space $\mathcal{A}$.} Our robot is a holonomic robot, and its actions are drawn from a continuous action space. The robot’s action $a_t$ can be defined as:
\begin{align} 
    a_t=[v_x,v_y] \in \mathbb{R}^2 .
\end{align}

\textbf{Reward Function $\mathcal{R}$}. Similar to \citet{liu2021decentralized}, during navigation, the robot incurs a negative reward when it enters the pedestrian danger zone or collides with pedestrians. Once the robot reaches the goal, it receives a positive reward. Additionally, the robot is encouraged to move towards the goal. After a time step, if the robot gets closer (farther away) from the goal, it receives a positive (negative) reward. Thus, the reward function $r_t(s_t, a_t)$ is defined as:
\begin{equation}\label{rewardf}
	r_t(s_t, a_t) = \begin{cases}
	-0.25,   &if \ d_{min}^t \leqslant 0\\
	d_{min} - 0.2,    &else \ if\  d_{min}^t < 0.2  \\
        1,    &else \ if \ d_g^t \leqslant r_{robot}  \\
        d_g^{t-1} - d_g^t,        &otherwise
		   \end{cases}
\end{equation}
where $d_{min}^t$ is the distance between the robot and the nearest pedestrian, and $d_g^t$ is the distance between the goal and the robot at time $t$. 

The objective of the robot is to find an optimal policy that maximizes state-action value function $Q^*(s_t, a_t)$:
\begin{align}
    \pi^* (s_t)= \underset{a_t}{\mathrm{arg\,max}}Q^*(s_t, a_t),
\end{align}
then $Q^*(s_t, a_t)$ is established by the Bellman optimal equaltion:
\begin{align}
    Q^*(s_t,a_t)= &\sum_{s_t',r_t}P(s_t',r_t \mid s_t,a_t) [r_t+\gamma \max_{a_t'}Q^*(s_t',a_t')].
\end{align}

\begin{figure}[t]
\centerline{\includegraphics[width=8.8cm, height=4cm]{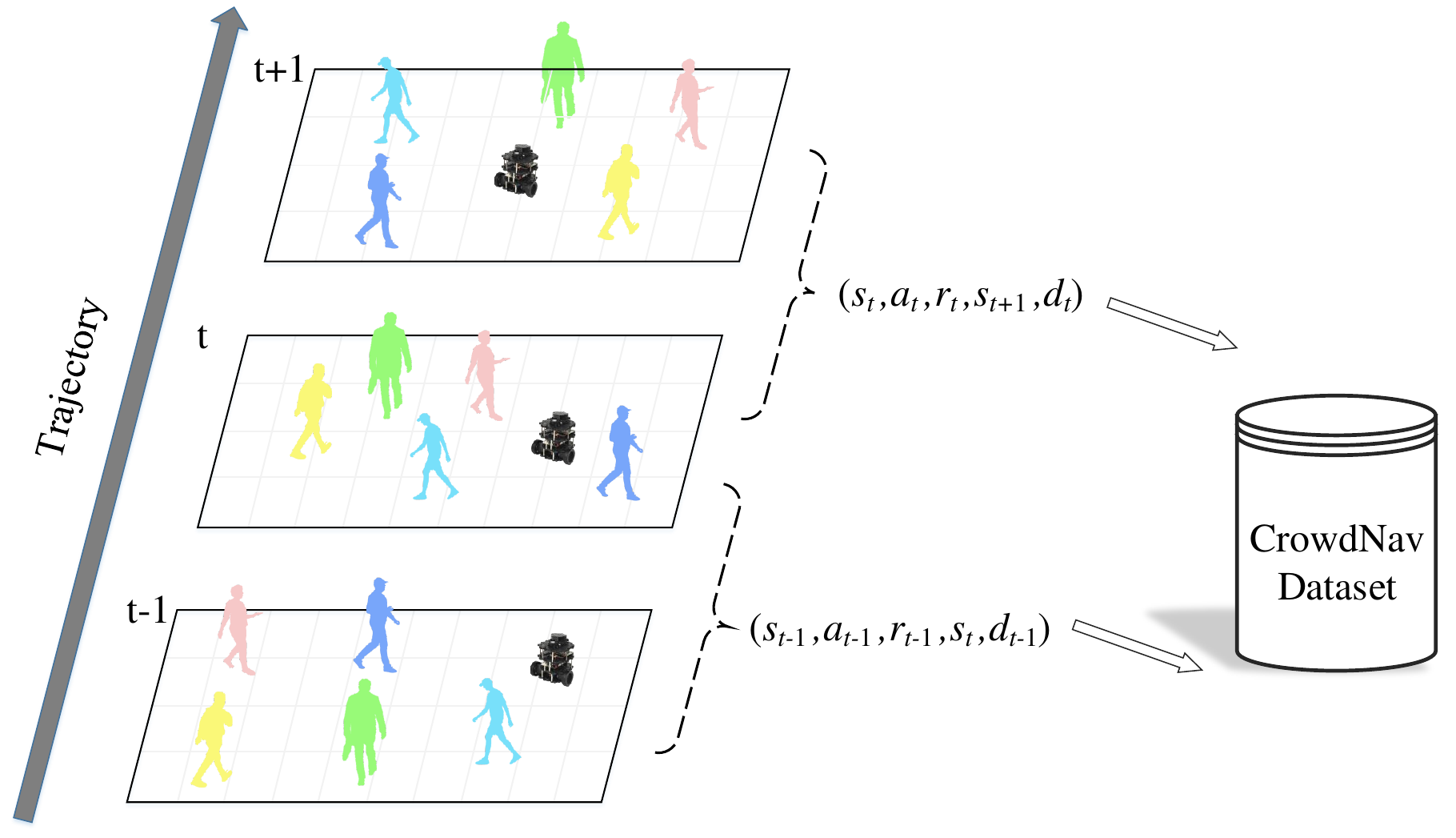}}
\caption{Demonstration of Robot Crowd Navigation Data Collection: In each round, the robot and pedestrians execute ORCA strategies from their initial positions. At each time step, the global state $s_t$, the action $a$ taken by the robot, the reward $r_t$ provided by the environment, the next global state $s_{t+1}$, and termination information $d_t$ are collected as a transition. As soon as the environment ends, a new environment begins, and this process continues until the dataset is fully collected.}
\label{env}
\end{figure}

\subsection{Offline Reinforcement Learning for Robot Crowd Navigation}
Currently, robot crowd navigation algorithms are primarily based on an online training mode, requiring extensive interaction with the environment to ensure the quality of learning
Instead, this paper proposes an offline reinforcement learning based robot crowd navigation algorithm. Specifically, implicit Q-Learning (IQL) \citep{kostrikov2021offline} is developed for robot crowd navigation. By modifying the loss function in the SARSA-style objective, we eliminate the querying of out-of-distribution robot actions from the pre-collected crowd navigation experience. Furthermore, this work also provides a data collection scheme for offline crowd navigation datasets to support our algorithm.
 
\textbf{Offline Learning}. During the value update phase, to eliminate queries for actions that out-of-distribution range of offline crowd navigation experiences, the state-action distribution from $\mathcal{D}$ is utilized to approximate the optimal Q-function $Q_\theta(s,a)$ through SARSA-style objective:
\begin{equation}\label{Q}
L_Q(\theta) = \mathbb{E}_{(s,a,s')\sim \mathcal{D}}[(r(s,a) + \gamma ^{\Delta t \cdot v_{pre}} V_\varphi(s') - Q_\theta(s,a))^2],
\end{equation}
where $\mathcal{D}$ is a crowd navigation dataset, $\Delta t$ is the time interval, and here $\Delta t$ = 0.25. $V_\varphi(s)$ is a separate value function that is trained to maximize the value corresponding to the optimal action for $Q_\theta(s, a)$ and is used to aid in Q-function learning. Its optimization objective is given by:
\begin{equation}\label{V}
L_V(\varphi) = \mathbb{E}_{(s,a)\sim \mathcal{D}}[L_2^\tau(Q_{\hat{\theta}}(s,a)-V_\varphi(s))],
\end{equation}
where $L_2^\tau(u) = |\tau - \mathbbm{1}(u \textless 0)|u^2$ is expected regression, and $Q_{\hat{\theta}}(s,a)$ is the target Q-network. 

During the policy update phase, the policy $\pi(\phi)$ is optimized through advantage weighted regression (AWR) \citep{peng2019advantage} to minimize the following objective:
\begin{equation}\label{p}
L_\pi(\phi) = \mathbb{E}_{(s,a)\sim \mathcal{D}}[\mathrm{exp}(\beta(Q_{\hat{\theta}}(s,a) - V_\varphi(s)) \mathrm{log} \pi_{\phi}(a|s)],
\end{equation}
where $\beta \in[0, \infty)$ is an inverse temperature.
For smaller hyperparameters, the target policy is similar to behavioral cloning. In contrast, for larger hyperparameters, it attempts to maximize the value of the Q-function.

\begin{algorithm}[t]
\caption{Data Preparation} \label{alg: A}  
\begin{algorithmic}[1]
\STATE Initialize a crowd environment
\STATE Initialize the empty dataset $\mathcal{D}$
\FOR {episode = 1 to N}
\STATE Initialize random state $s_0$ and set $t = 0$
\WHILE{no Reach goal, Collision or Timeout}
\STATE Select action with exploration noise:
\STATE $a_t \leftarrow ORCA(s_t) + \epsilon , \epsilon \sim \mathcal{N}(0, 0.1) $
\STATE Execute $a_t$ and obtain $r_t ,s_{t+1}$ and $d_{t}$
\STATE Store experiences 
\STATE $\mathcal{D} \leftarrow \mathcal{D} \cup \{(s_t, a_t, r_t, s_{t+1}, d_{t}) \}$
\STATE $s_t \leftarrow s_{t+1}$
\IF {$\mathcal{D}$ is full}
\STATE break

\ENDIF
\ENDWHILE
\ENDFOR
\ENSURE{the dataset $\mathcal{D}$}
\end{algorithmic}  
\end{algorithm}

\textbf{ Crowd Navigation Dataset.}\label{Data Preparation} Due to the scarcity of general datasets for robot crowd navigation, acquiring an efficient crowd navigation dataset is necessary for our work. As shown in Fig.\ref{env}, this work follows Section \ref{modeling} modeling to collect the dataset $\mathcal{D}$. It is represented as:
\begin{align}
    \mathcal{D} = \{ (s_i, a_i, r_i, s'_i, d_i) | i=1, 2,...,N \}, 
\end{align}
where $(s_i, a_i, r_i, s'_i, d_i)$ denotes the $i$-th transition in the dataset. $d$ represents environment termination information, and $N$ represents the maximum capacity of the dataset. 

Our dataset collected for this task utilizes ORCA \citep{van2011reciprocal} as the controlling policy for the robot. Furthermore, to enhance dataset robustness and encourage exploration, the robot's sampled actions are augmented with exploratory noise drawn from $\mathcal{N}(0, 0.1)$. Algorithm \ref{alg: A} illustrates the detailed procedure.

Similar to D4RL \citep{fu2020d4rl}, the dataset is stored in Hierarchical Data Format (HDF5), a file format used for storing scientific data. For more detail about our dataset, see Section \ref{Dateset01}.

\begin{algorithm}[t]
\caption{Network Training} \label{alg: B}  
\begin{algorithmic}[1]
\STATE Initialize Q-network $Q_\theta$, value network $V_\varphi$ and policy network $\pi_\phi$ with random parameters $\theta, \varphi $ and $\phi$
\STATE Initialize target Q-network $ Q_{\hat{\theta}}$ with parameters $ \hat{\theta} \leftarrow \theta$
\STATE Initialize the replay buffer $\mathcal{B}$
\STATE Load all transitions from $\mathcal{D}$ into  replay buffer $\mathcal{B} = \{ (s^{jn}_i, a_i, s^{jn'}_i ,r_i, d_i) | i=1, 2,...,N \}$
\FOR {iteration i=1 to M}
\STATE Sample m random transitions $(s^{jn}, a, r, s^{jn'}, d) \thicksim \mathcal{B}$
\STATE $y \leftarrow r + (1-d)\gamma V_\varphi(s^{jn'})$
\STATE $ L_Q(\theta) \leftarrow m^{-1} \sum \frac{1}{2}(y -Q_\theta(s^{jn}, a))^2$
\STATE Updata Q-network $Q_\theta$ by using loss function $L_Q(\theta)$
\STATE Update target Q-network $ Q_{\hat{\theta}}$: $\hat{\theta} \leftarrow (1 - \alpha) \hat{\theta} + \alpha \theta$
\STATE$ adv \leftarrow Q_{\hat{\theta}}(s^{jn},a)-V_\varphi(s^{jn})$
\STATE $L_V(\varphi) \leftarrow  m^{-1} \sum L_2^\tau(adv)$ 
\STATE Updata value network $V_\varphi$ by using loss function $L_V(\varphi)$

\STATE $L_\pi(\phi) \leftarrow m^{-1} \sum \mathrm{exp}(\beta \ adv) \mathrm{log} \pi_{\phi}(a|s)$
\STATE Update policy network $\pi_\phi$ by using loss function $L_\pi(\phi)$

\ENDFOR
\end{algorithmic}  
\end{algorithm}

\begin{figure*}[t]
\centerline{\includegraphics[width=0.99\textwidth]{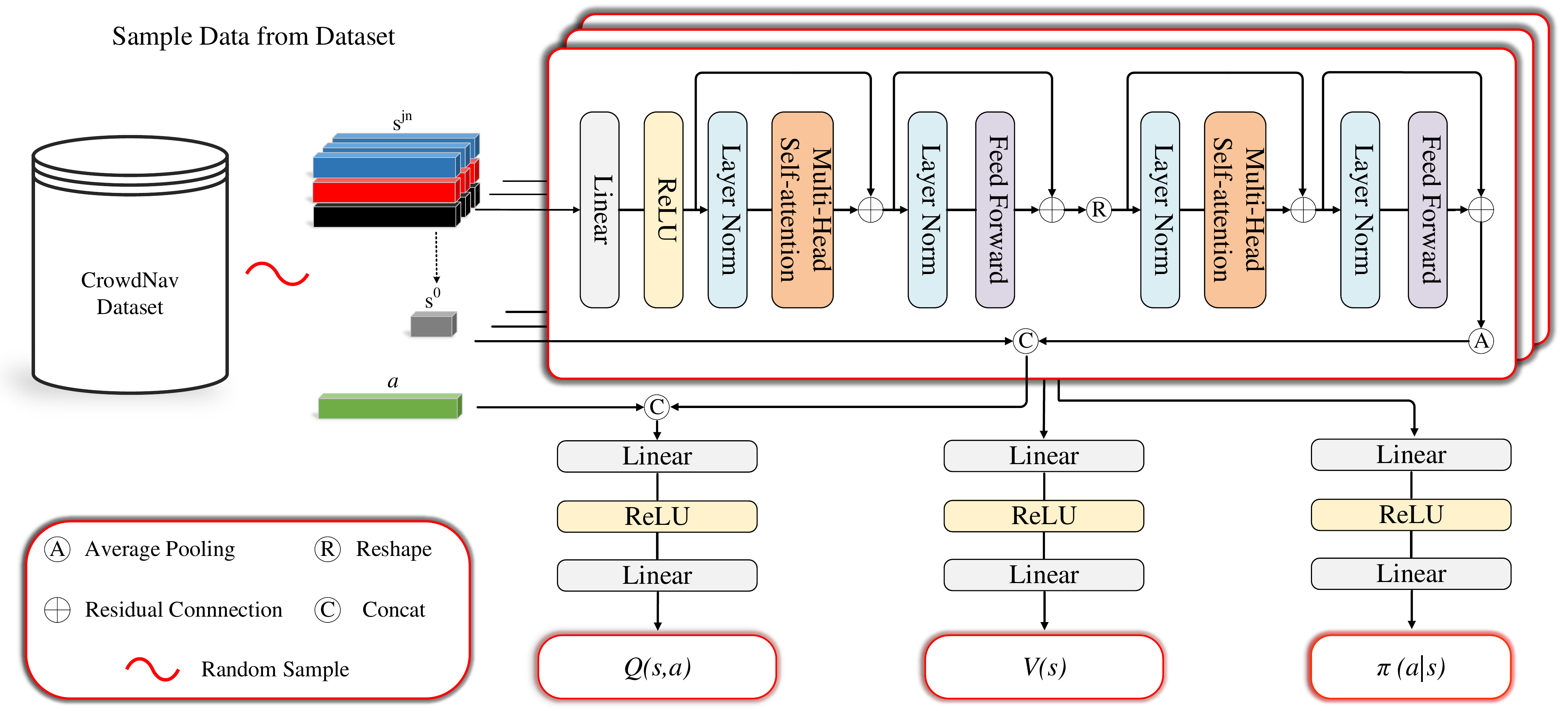}}
\caption{The architecture of ST2-ORL. This architecture consists of three separate networks: the Q-network, the value network, and the policy network. The joint state $s^{jn}$ sampled from the crowd navigation dataset is fed into the spatial-temporal state transformer for encoding. Then, the encoded state is combined with the robot's state $s^0$ to form latent features. Finally, the Q-network takes latent features with the actions $a$ as input and outputs Q values, the value network takes latent features as input and outputs V values, while the policy network takes latent features as input and outputs target policy.}
\label{framework}
\end{figure*}

\subsection{Network Architecture}\label{AA}
In robot crowd navigation, the complexity of the crowd environment makes it challenging for the low-generalization global state $s$ to adequately fit the tasks of $Q$-value, $V$-value, and $\pi$. To explore the potential spatial-temporal features in the global state $s$, the spatial-temporal state transformer is integrated into IQL to encode the pedestrian-robot interactions. Specifically, ST2-ORL is proposed, which takes the joint state with spatial-temporal relationships from the dataset $\mathcal{D}$ as input. It then extracts effective spatial-temporal representations for IQL through the global spatial state encoder and the local temporal state encoder. Afterward, it follows algorithm \ref{alg: B} to train a planner.

The model takes the spatial-temporal joint state $s^{jn} \in \mathbb{R}^{3\times H\times L}$ as input, which is formed by concatenating the global state from 3 consecutive time steps along the temporal dimension: 
\begin{equation}
    s^{jn} = \begin{cases}
    [s_0, s_0, s_0], & if \ t = 0\\
    [s_0, s_0, s_1], & else \ if \ t = 1\\
    [s_{t-2}, s_{t-1}, s_t], & otherwise
    \end{cases}
\end{equation}
where $H$ represents the number of pedestrians, and $L$ represents the sequence length of the joint state.

Then, $s^{jn}$ is embedded into higher dimensions for preliminary feature extraction. The feature $f \in \mathbb{R}^{3\times H\times \hat{L}}$ is represented as:
\begin{equation}
    f = f_p(s^{jn};W_p),
\end{equation}
where $f_p$ is a fully connected layer with the ReLU activation, and $W_p$ represents the weight parameters.

The global spatial state encoder views the time dimension as the batch dimension and captures the spatial positional relationships between the robot and different pedestrians. This encoder emphasizes the importance of different pedestrians to the robot through the spatial multi-head self-attention layer (Spatial-MSA). Then a feedforward neural network (FNN) maps the spatial position relationship to a higher-dimensional feature space. It takes the preliminary extracted feature $f$ as input and outputs enhanced features with spatial dependencies. The global spatial encoder equation is formulated as below:
\begin{align}
    Q_s &= f_{qs}(f; W_{qs}),\\
    K_s &= f_{ks}(f; W_{ks}), \\
    V_s &= f_{vs}(f; W_{vs}),
\end{align}
where $f_{qs}, f_{ks} \ and \ f_{vs} $ are fully connected layers with the ReLU activation,
 respectively, $W_{qs}, W_{ks} \ and \ W_{vs}$ represents the weight parameters, and $Q_s, K_s \ and \ V_s$ are the query, key and value vectors. Then, the multi-head self-attention mechanism is used to capture spatial dependencies by concluding the attention scores for each pedestrian with respect to the robot.
 \begin{equation}\label{equ16}
     Att_i(Q_s, K_s, V_s) = softmax(\dfrac{Q_s K_s^T}{\sqrt{d_k}})V_s,
\end{equation}
\begin{equation}
    head_i = Att_i(Q_s, K_s, V_s),
\end{equation}
\begin{equation}
    Spatial-MSA(Q_s, K_s, V_s) = f_{o}([head_i]^h_{i=1}),
\end{equation}
where $Att_i(Q_s, K_s, V_s)$ is a self-attention head, $f_o$ is a fully connected layer that merges the h heads, $d_k$ is the dimension of $Q$ and $K$. The multi-head
attention layer’s output is then fed into FNN via residual connections and normalization layers:
\begin{equation}
    f_s' = Spatial-MSA(Q_s, K_s, V_s) + f,
\end{equation}
\begin{equation}\label{equ20}
    f_s = FNN(LN(f_s')) + f_s',
\end{equation}
where FNN refers to the two-layer fully connected with the ReLU activation.

The pedestrian-robot interactions in the time dimension also play an important role in the robot's decision-making.
The local temporal state encoder captures the temporal dynamic evolution clues of each pedestrian through a temporal multi-head self-attention layer (Temporal-MSA). Similarly, a FNN is used to map the temporal dynamic transformation relationships to a higher-dimensional feature space. 

To emphasize the temporal dimension, it first transposes the spatial enhanced features $f_s$ to $f_s^T \in \mathbb{R}^{H\times 3\times \hat{L}}$ and then feeds it as input. It outputs the comprehensive spatial-temporal enhanced feature $f_{st} \in \mathbb{R}^{H \times 3\times \hat{L}}$. In the local temporal state encoder, the query, key and value vector are rewritten as follows:
\begin{align}
    Q_t &= f_{qt}(f_s^T; W_{qt}),\\
    K_t &= f_{kt}(f_s^T; W_{kt}), \\
    V_t &= f_{vt}(f_s^T; W_{vt}).
\end{align}
where $f_{qt}, f_{kt} \ and \ f_{vt} $ are fully connected layers with the ReLU activation,
 respectively, $W_{qt}, W_{kt} \ and \ W_{vt}$ represents the weight parameters. $Q_t, K_t \ and \ V_t$ are the query, key and value vectors. And the Equation (\ref{equ16}) – (\ref{equ20}) are rewritten as follows:
 \begin{equation}
     Att_i(Q_t, K_t, V_t) = softmax(\dfrac{Q_t K_t^T}{\sqrt{d_k}})V_t,
\end{equation}
\begin{equation}
    head_i = Att_i(Q_t, K_t, V_t),
\end{equation}
\begin{equation}
    Temporal-MSA(Q_t, K_t, V_t) = f_0([head_i]^h_{i=1}),
\end{equation}
\begin{equation}
    f_{st}' = Temporal-MSA(Q_t, K_t, V_t) + f_t^T,
\end{equation}
\begin{equation}
    f_{st} = FNN(LN(f_{st}')) + f_{st}',
\end{equation} 

Finally, spatial-temporal enhanced feature $f_{st}$ is fed into 3 individual networks to fit the $Q$-value, $V$-value and $\pi$:
\begin{align}
    Q &= f_\theta([Average(f_{st}^1), s^r_t, a_t];W_\theta),\\
    V &= f_\varphi([Average(f_{st}^2), s^r_t];W_\varphi),\\
    \pi &= f_\phi([Average(f_{st}^3), s^r_t]);W_\phi),
\end{align}
where $f_\theta$ , $f_\varphi$ and $f_\phi$ are two-layer fully connected with Relu activation, $W_\theta$ , $W_\varphi$ and $W_\phi$ are the weight matrices of the three networks. $f_{st}^*$ represents the enhanced features generated by the spatial-temporal transformer in each individual network. $\pi$ is a multidimensional distribution based on action, used for sampling actions.
Fig. \ref{framework} shows the proposed ST2-ORL network structure.

\section{EXPERIMENTS}\label{4}
\subsection{Experiment Setup}
\textbf{Simulation Environment.} The experiments are conducted within the CrowdNav environment, as explained in Section \ref{modeling}. Specifically, we design simple scenarios and complex scenarios, respectively. In simple scenarios, 6 pedestrians are randomly positioned within a circular area with a 4-meter radius. For each pedestrian, their initial position and initial target position are set on the circumference of the circle, symmetrically around the center. In addition to circular crossing pedestrians, the complex scenario introduces square crossing pedestrians. These pedestrians are randomly placed within a square area with sides of 10 meters. And this square is aligned with the centerline of the circle. A total of 9 pedestrians are randomly selected to be either circular or square crossing pedestrians. Pedestrians employ the ORCA \citep{van2011reciprocal} strategy for navigation and obstacle avoidance. Regardless of whether it is in simple scenarios or complex scenarios, pedestrians do not have a final destination. Instead, each pedestrian immediately proceeds to a newly generated random goal upon reaching their current one.

\begin{table}[htbp]
\setlength{\abovecaptionskip}{3pt}
\caption{Specific performance metrics of datasets}
\label{dataset}
\begin{center}
\begin{tabular}{cccccc}
   \toprule
   Datasets & Success & Collision & Time & Reward & Capacity\\
   \midrule
   simple & 92.7\% & 7.3\% & 14.0s & 0.779 & $5 \times 10^5$  \\
   complex & 91.6\% & 8.3\% & 16.2s & 0.640 & $5 \times 10^5$   \\
   \bottomrule
\end{tabular}
\end{center}
\end{table}

\textbf{Creating Datasets.}\label{Dateset01} 
Our datasets are built within this simulation environment. In our setup, the robot is configured to be invisible. Unfortunately, in the case of invisibility, especially in complex scenarios, robots are prone to collisions with pedestrians. To address this, safety space are incorporated into the robot's policy, ensuring that the robots can successfully avoid pedestrians to some extent. Here, set the safety space of the robot to 0.2 to ensure the quality of navigation data.

We collected data in simple and complex scenarios respectively, and two datasets named ``simple" and ``complex" were created accordingly. Details of these datasets, five metrics, i.e., ``$Success$", ``$Collision$", ``$Time$", ``$Reward$" and ``$Capacity$" are shown in Table \ref{dataset}. They describe the success rate, collision rate, average navigation time of success trajectories, average cumulative reward of all trajectories and datasets capacity, respectively. From the performance perspective, both datasets are suboptimal. 

\textbf{Baseline.} 6 state-of-the-art algorithms. ORCA as the baseline for reaction-based method; Behavior Cloning (BC) as the baseline for imitation learning;
IQL \citep{kostrikov2021offline} as the baseline for offline reinforcement learning; CADRL \citep{chen2017decentralized}, LSTM-RL \citep{everett2018motion}, SARL \citep{chen2019crowd} and DS-RNN \citep{liu2021decentralized} as the baselines for robot crowd navigation based on traditional DRL.

\textbf{Training Settings}. In our algorithm, each individual network utilizes the Adam optimizer \citep{kingma2014adam} with a learning rate of 0.0005 and is trained with 128 batches. The discount factor $\gamma$ is set to 0.9, the inverse temperature $\beta$ is set to 100, and the expected regression factor $\tau$ is set to 0.8.

In the implementation of ORCA, the robot's safety space is set to 0.2, consistent with the policy in the dataset. Both BC and IQL use the same datasets and train with the same training parameters as our algorithm. CADRL, LSTM-RL, SARL and DS-RNN employ the same rewards as defined in Equation \ref{rewardf}. And their network architectures and training settings are consistent with the original papers.

\subsection{Quantitative Evaluation}
During the testing process, all methods are evaluated for performance in 500 testing environments, consisting of both simple and complex scenarios.
In this experiment, ``$Success$", ``$Collision$", ``$Time$" and ``$Reward$" are used to evaluate performance. Additionally, to discuss performance related to sampling efficiency, ``$Efficiency$" metric is listed to quantify the sampling efficiency of all methods. The ``$Efficiency$" is defined as:
\begin{align}
    Efficiency = average \  cumulative \  reward \ / \ total \ transitions * 10^5.
\end{align}
Note that, this field is empty for the ORCA. The performance under the two different scenarios is presented in Table \ref{simplet} and \ref{complext}.

\begin{table}[t]
\caption{Quantitative results of all methods in simple scenarios}
\label{simplet}
\begin{center}
\begin{tabular}{cccccc}
   \toprule
   Methods & Success & Collision & Time & Reward & Efficiency\\
   \midrule
    ORCA & 93.4\% & 6.6\% & 14.0s & 0.785 & -\\  
    \midrule
    CADRL & 95.2\% & 4.2\% & 17.2s & 0.560 & 0.067\\  
    LSTM-RL & 96.6\% & 3.8\% & 13.2s & 0.788 & 0.107\\ 
    DS-RNN & 95.6\% & 4.4\% & \textbf{12.0s} & 0.905 & 0.090   \\ 
    SARL & \textbf{99.4}\% & \textbf{0.6\%} & \textbf{11.4s} & \textbf{1.025} &0.147\\ 
    \midrule
    BC & 85.6\% & 14.4\% & 14.0s & 0.707  & 0.141\\ 
    IQL & 89.4\% & 10.6\% & 12.2s & 0.786 & 0.157 \\ 
    ST2-ORL & \textbf{98.4\%} & \textbf{1.6\%} & 12.5s & \textbf{0.960} &  \textbf{\text{0.192}} \\ 
   \bottomrule
\end{tabular}
\end{center}
\end{table}

\begin{table}[t]
\caption{Quantitative results of all methods in complex scenarios}
\label{complext}
\begin{center}
\begin{tabular}{cccccc}
   \toprule
   Methods & Success & Collision & Time & Reward & Efficiency\\
   \midrule
    ORCA & 92.8\% & 7.2\% & 16.2s & 0.656 & -\\
    \midrule
    CADRL & 83.0\% & 12.2\% & 25.0s & 0.156 & 0.018\\ 
    LSTM-RL & 96.2\% & 3.4\% & 15.0s & 0.667 & 0.092\\ 
    DS-RNN & 93.4\% & 6.6\% & 14.0s & 0.721 & 0.072  \\ 
    SARL & \textbf{98.4}\% & \textbf{1.6\%} & \textbf{11.8s} & \textbf{0.971} &0.136\\ 
    \midrule
    BC & 80.2\% & 19.8\% & 18.2s & 0.464  & 0.093\\ 
    IQL & 86.0\% & 14.0\% & \textbf{13.4s} & 0.657 & 0.131 \\ 
    ST2-ORL & \textbf{97.6\%} & \textbf{2.4\%} & \textbf{13.5s} & \textbf{0.873} &  \textbf{0.175} \\ 
   \bottomrule
\end{tabular}
\end{center}
\end{table}

As shown in the first and last rows of the two tables, despite our experimental dataset samples from ORCA, our algorithm has achieved an improvement of approximately $5\%$ in success rate compared to ORCA, along with approximately $12\%$ reduction in navigation time.
The results indicate that our algorithm exhibits the capability to learn and infer excellent policies from suboptimal historical experiences in an offline mode through a SARSA-style objective.

It can be observed from the last three rows of the two tables that our algorithm exhibits more performance comprehensively. Unlike BC, which directly maps inputs to actions, resulting in a policy closely matching the dataset distribution, our algorithm learns long-term rewards in the SARSA-style objective, yielding a superior target policy. Although our algorithm lacks a slight advantage in average navigation time compared to the IQL baseline, it greatly outperforms the IQL by approximately $10\%$ in terms of success rate. This difference arises because IQL directly employs low-generalization pedestrian-robot interaction features, struggling with complex and dynamic crowd interactions, hindering success rate improvement. Conversely, our algorithm effectively captures spatial-temporal features from offline pedestrian-robot interaction experiences and guides efficient decision-making.

Compared to 4 traditional online learning baselines, our offline learning algorithm outperforms CADRL and LSTM-RL across all metrics. Moreover, the differences in success rate and average navigation time compared to DS-RNN and SARL are minimal. For instance, in simple scenarios, our algorithm's average navigation time is only slightly more than DS-RNN by less than 1 second. Even in complex scenarios, our algorithm lags behind SARL by only $0.8\%$ in success rate and $0.11\%$ in reward. The reason leading to these slight gaps is attributed to our offline learning algorithm lacking exploration capability, which hinders its ability to extract more effective pedestrian-robot interactions from training.

From the last column of the two tables, it is apparent that our algorithm's sampling efficiency surpasses SARL, the highest-performing one in traditional DRL baselines, by almost $30\%$ and almost $29\%$ respectively. This is because traditional online learning DRL baselines typically require frequent interaction with the environment to acquire more extensive learning experiences. The favorable results provide evidence that our algorithm, by combining IQL and spatial-temporal state transformer, effectively reduces exploration risk while enhancing sampling efficiency during pedestrian-robot interactions.

\begin{figure*}[t]
\centering
\subfigure[BC]{
\label{bcs} \includegraphics[width = 3.19cm, height = 3.19cm]{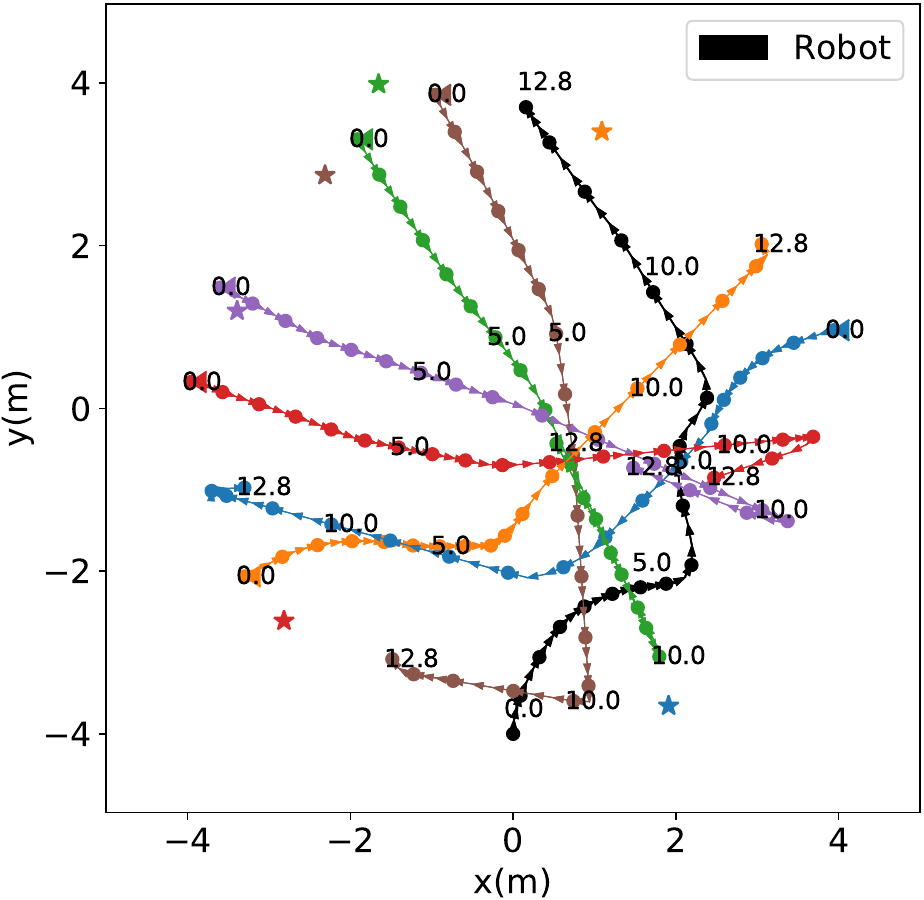}}
\subfigure[IQL]{
\label{iqls} \includegraphics[width = 3.19cm, height = 3.19cm]{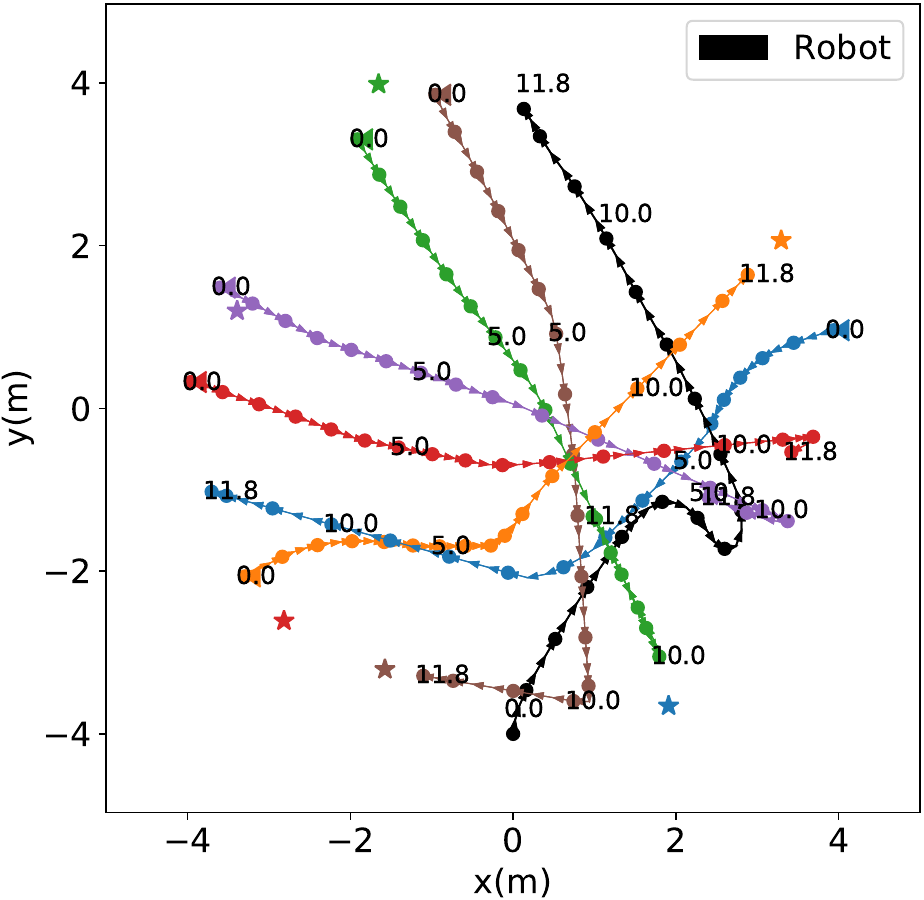}}
\subfigure[ORCA]{
\label{orcas} \includegraphics[width = 3.19cm, height = 3.19cm]{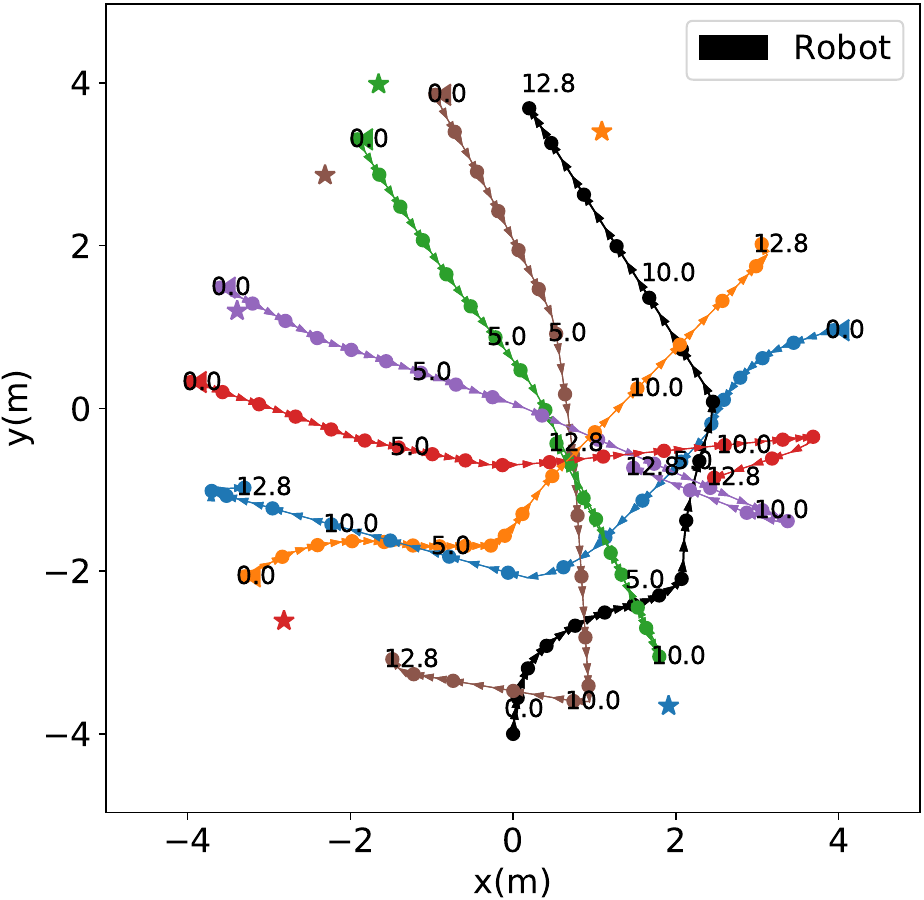}}
\subfigure[CADRL]{
\label{cadrls} \includegraphics[width = 3.19cm, height = 3.19cm]{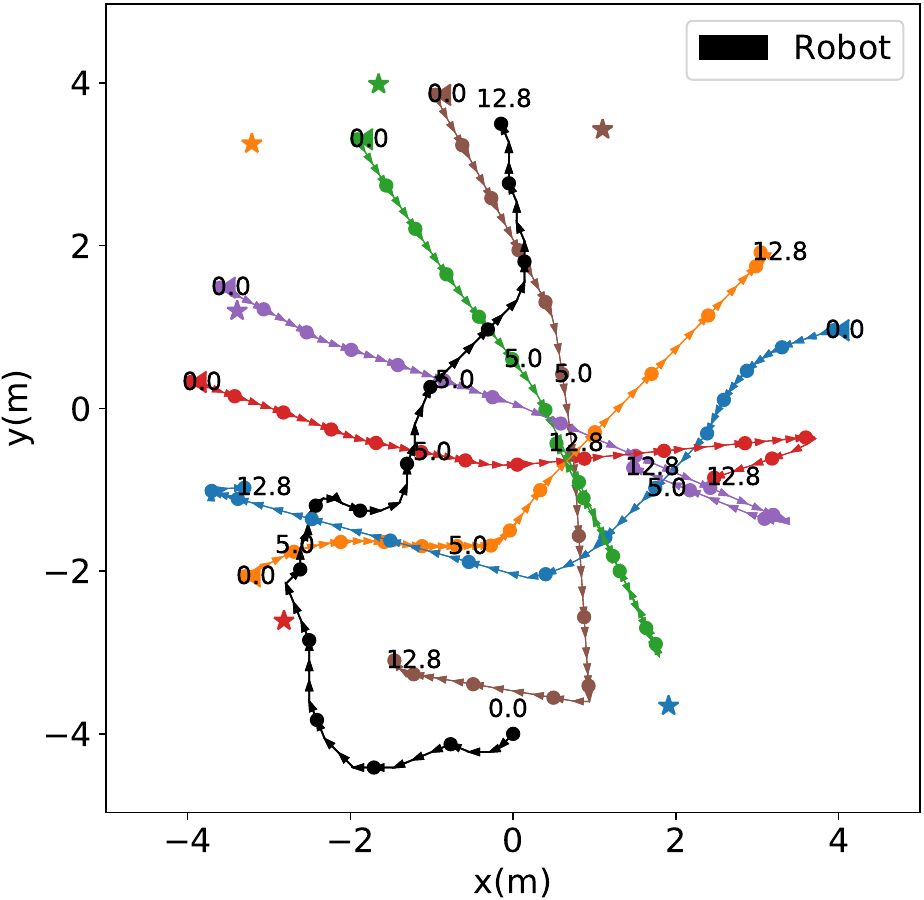}}
\subfigure[LSTM-RL]{
\label{lstms} \includegraphics[width = 3.19cm, height = 3.19cm]{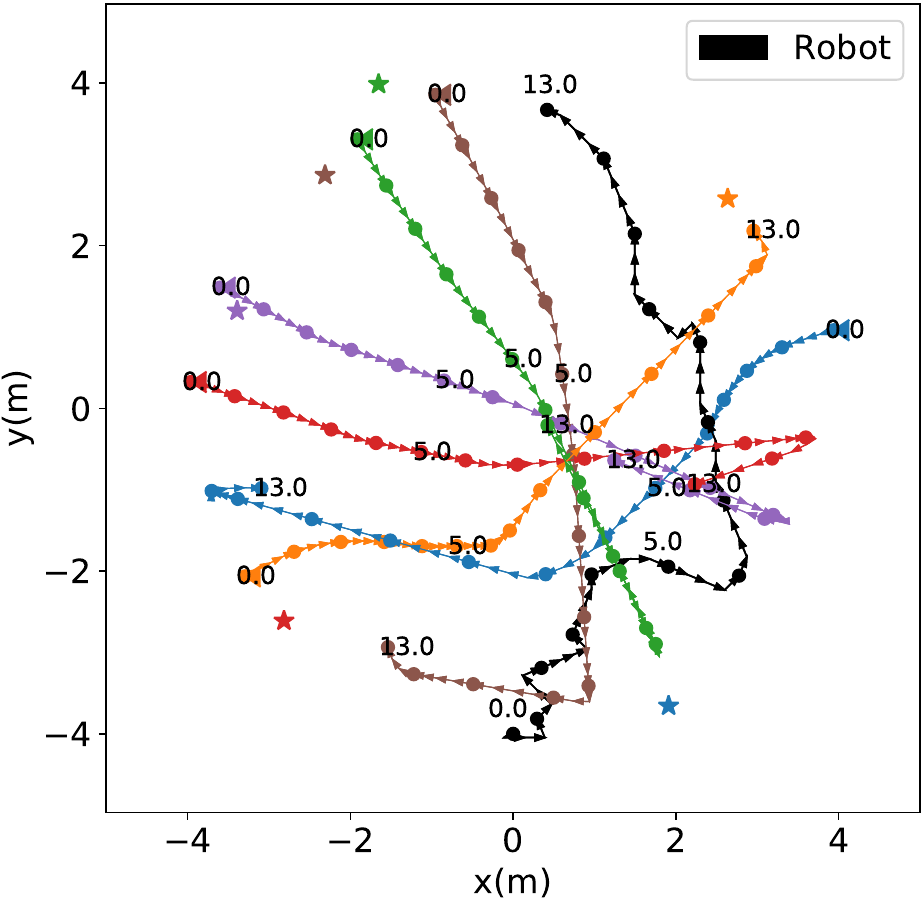}}
\subfigure[DS-RNN]{
\label{rnns} \includegraphics[width = 3.19cm, height = 3.19cm]{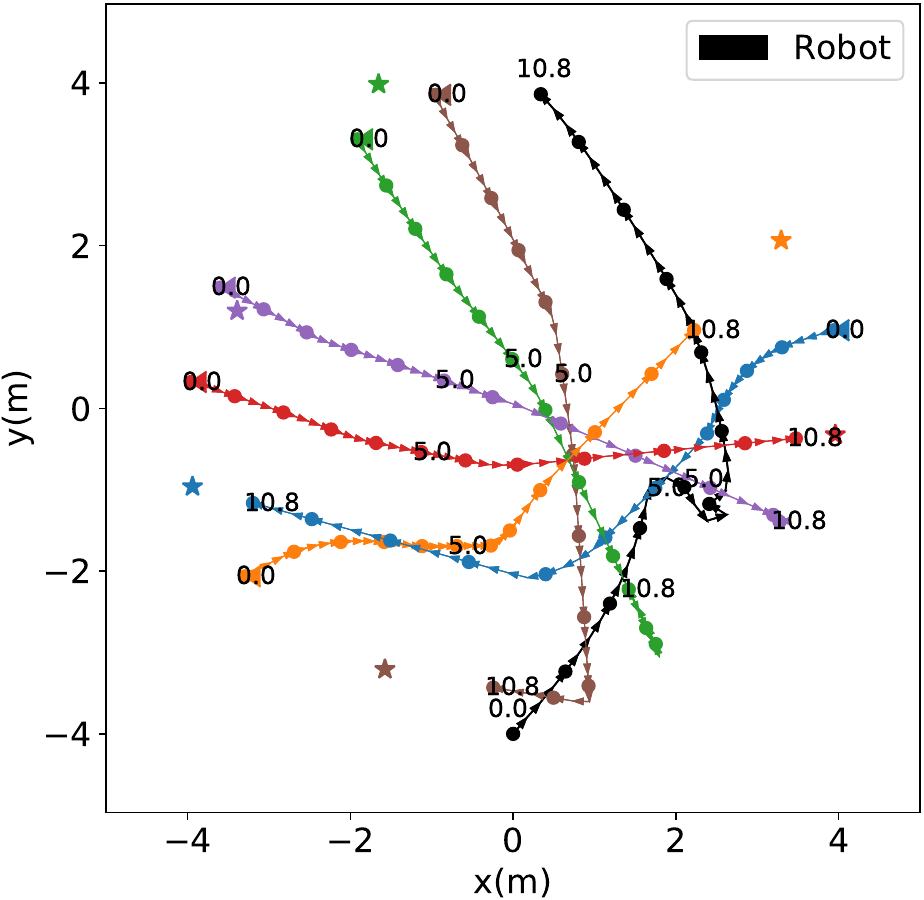}}
\subfigure[SARL]{
\label{sarls} \includegraphics[width = 3.19cm, height = 3.19cm]{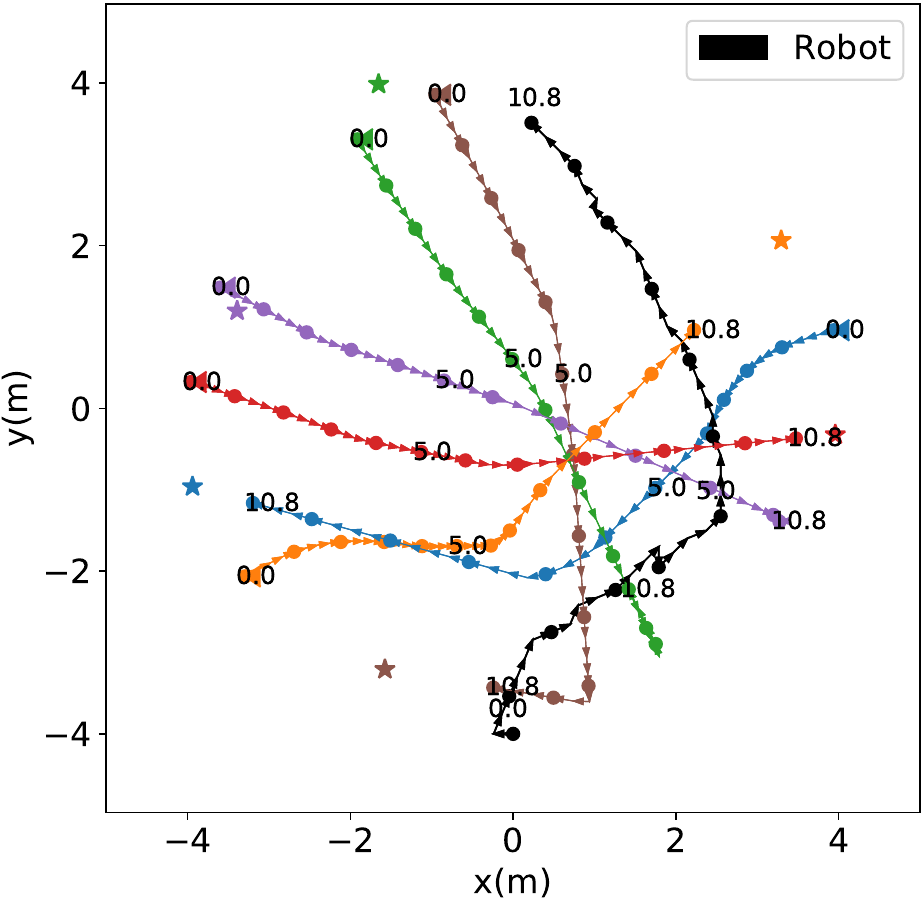}}
\subfigure[ST2-ORL]{
\label{st2s} \includegraphics[width = 3.19cm, height = 3.19cm]{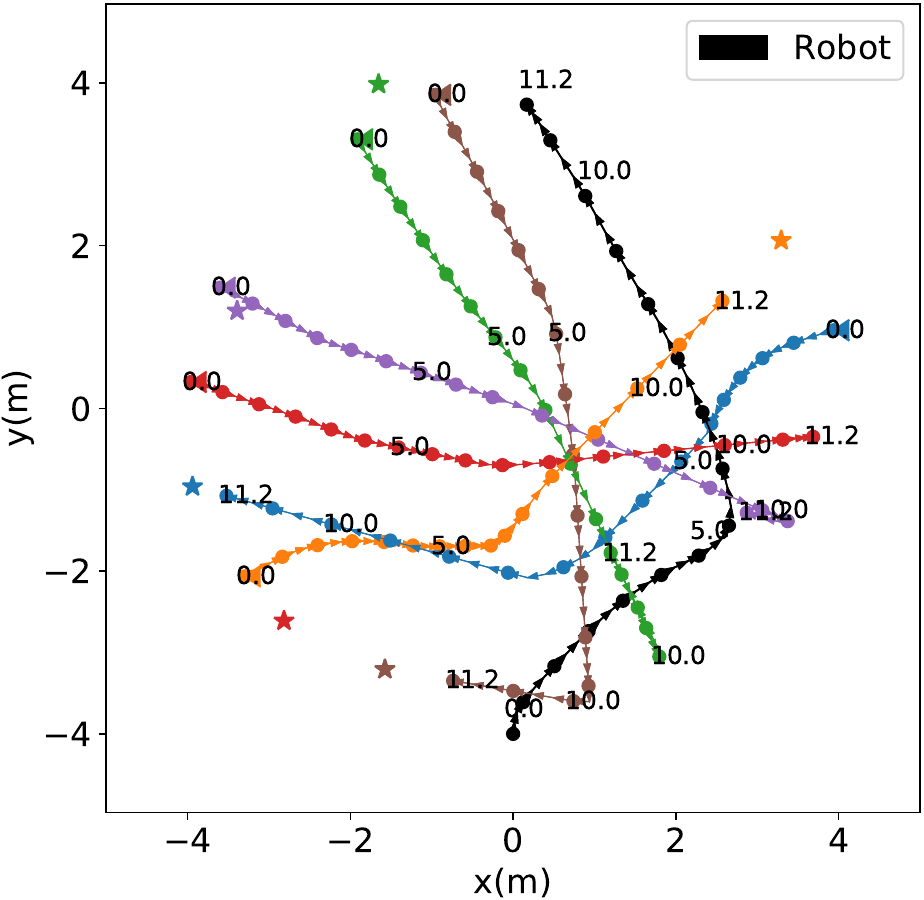}}
\caption{Illustration of trajectory in simple scenarios. Black indicates the trajectory of the robot, while other colors represent the trajectories of pedestrians.}
\label{simplep}
\end{figure*}

\begin{figure*}[t]
\centering
\subfigure[BC]{
\label{bcc} \includegraphics[width = 3.19cm, height = 3.19cm]{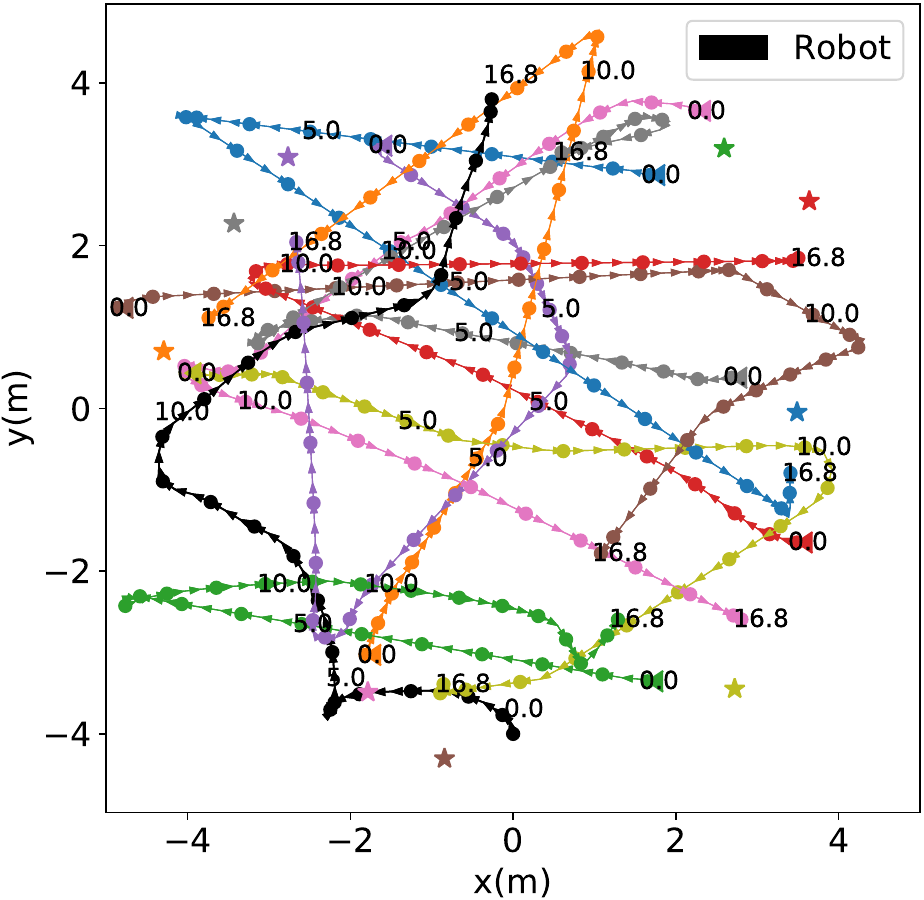}}
\subfigure[IQL]{
\label{iqlc} \includegraphics[width = 3.19cm, height = 3.19cm]{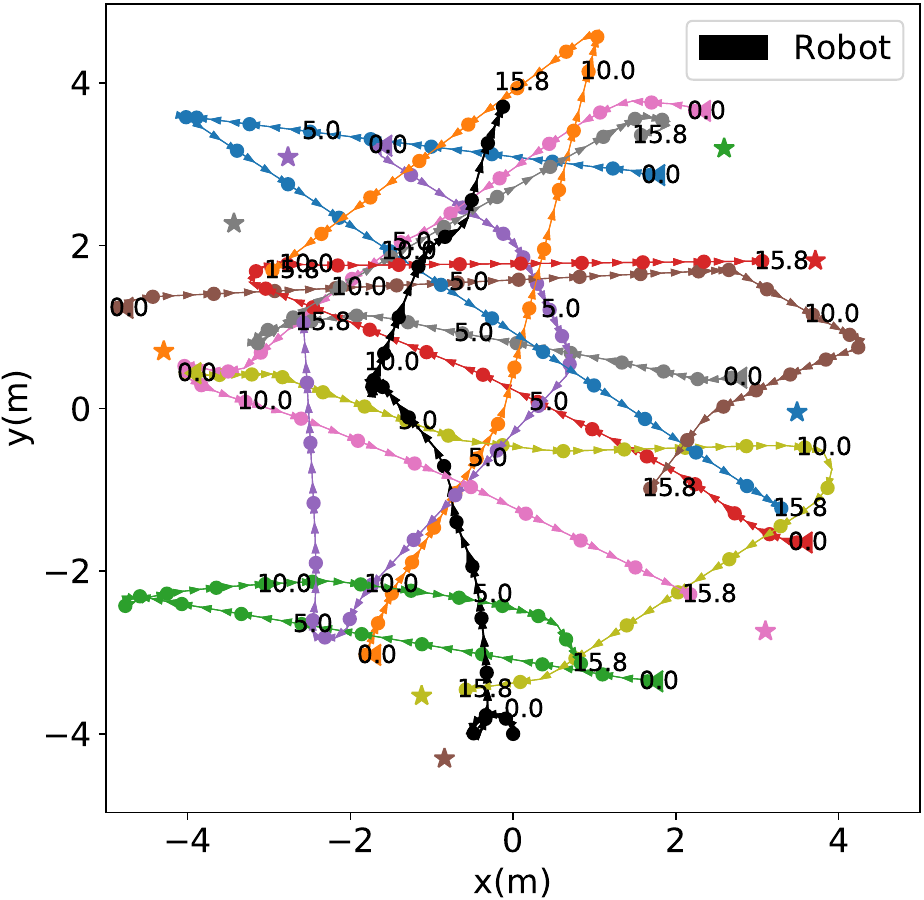}}
\subfigure[ORCA]{
\label{orcac} \includegraphics[width = 3.19cm, height = 3.19cm]{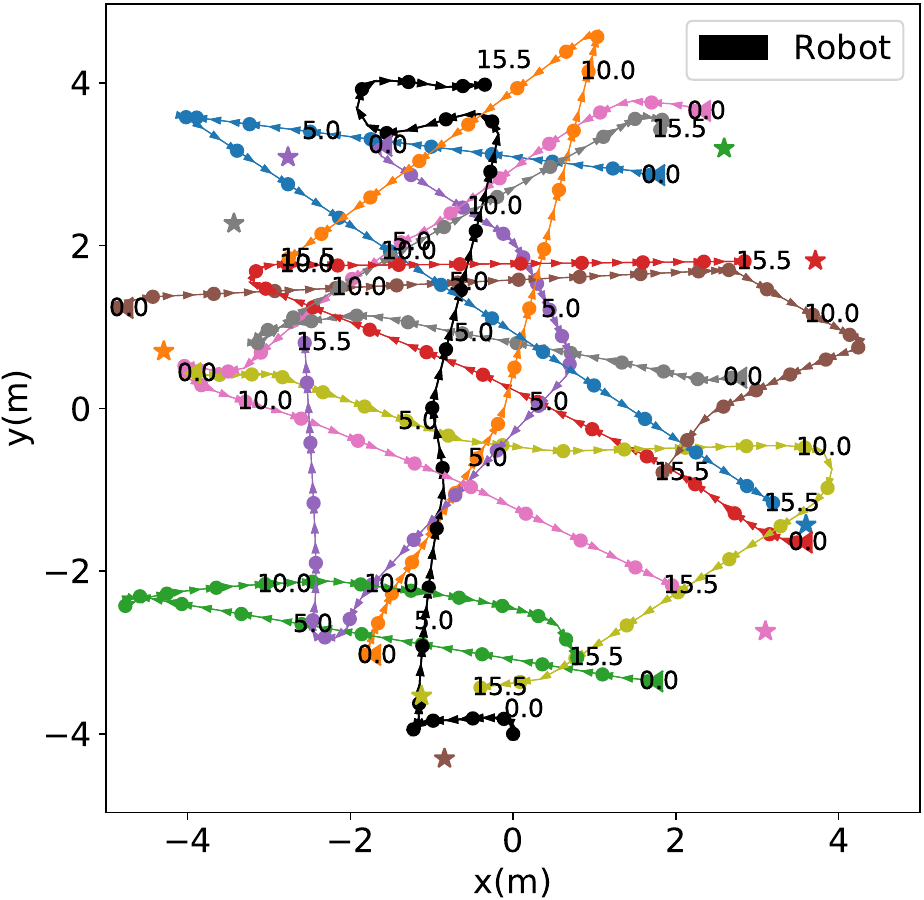}}
\subfigure[CADRL]{
\label{cadrlc} \includegraphics[width = 3.19cm, height = 3.19cm]{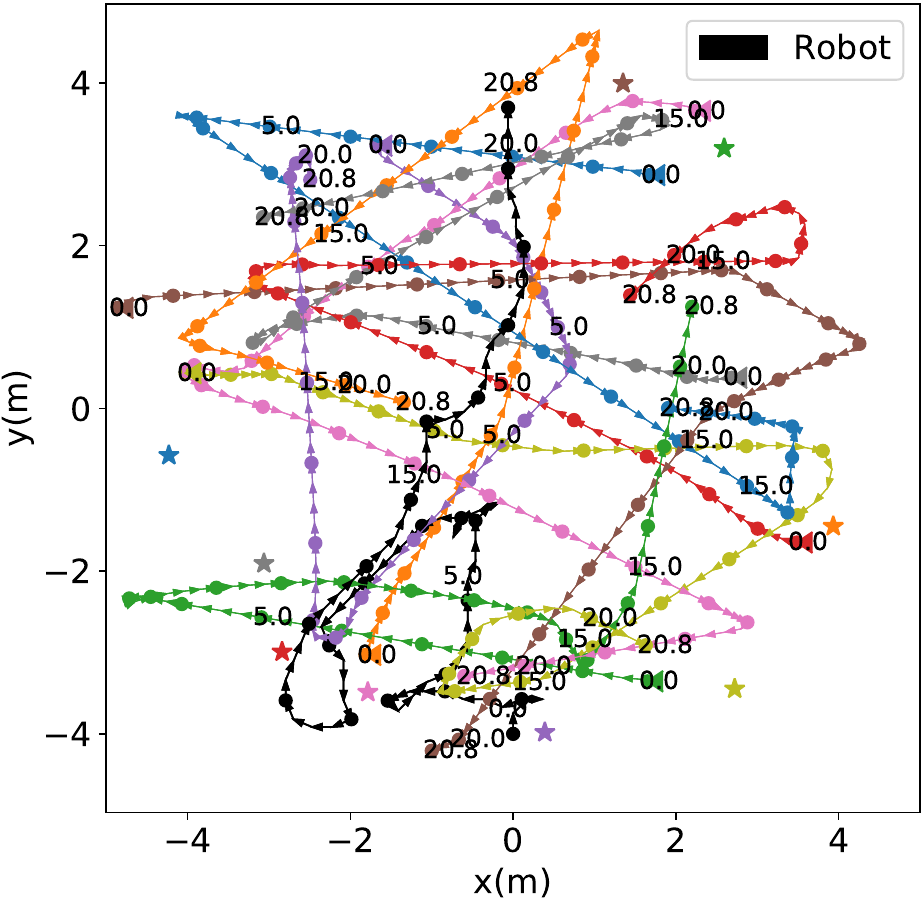}}
\subfigure[LSTM-RL]{
\label{lstmc} \includegraphics[width = 3.19cm, height = 3.19cm]{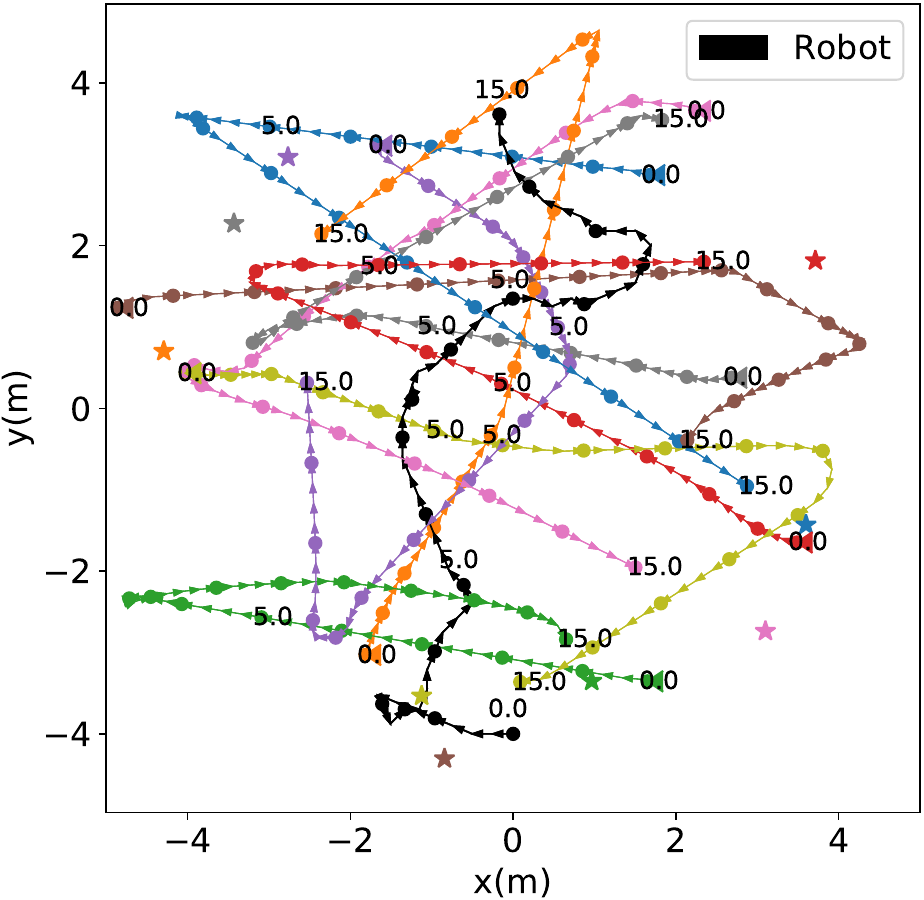}}
\subfigure[DS-RNN]{
\label{rnnc} \includegraphics[width = 3.19cm, height = 3.19cm]{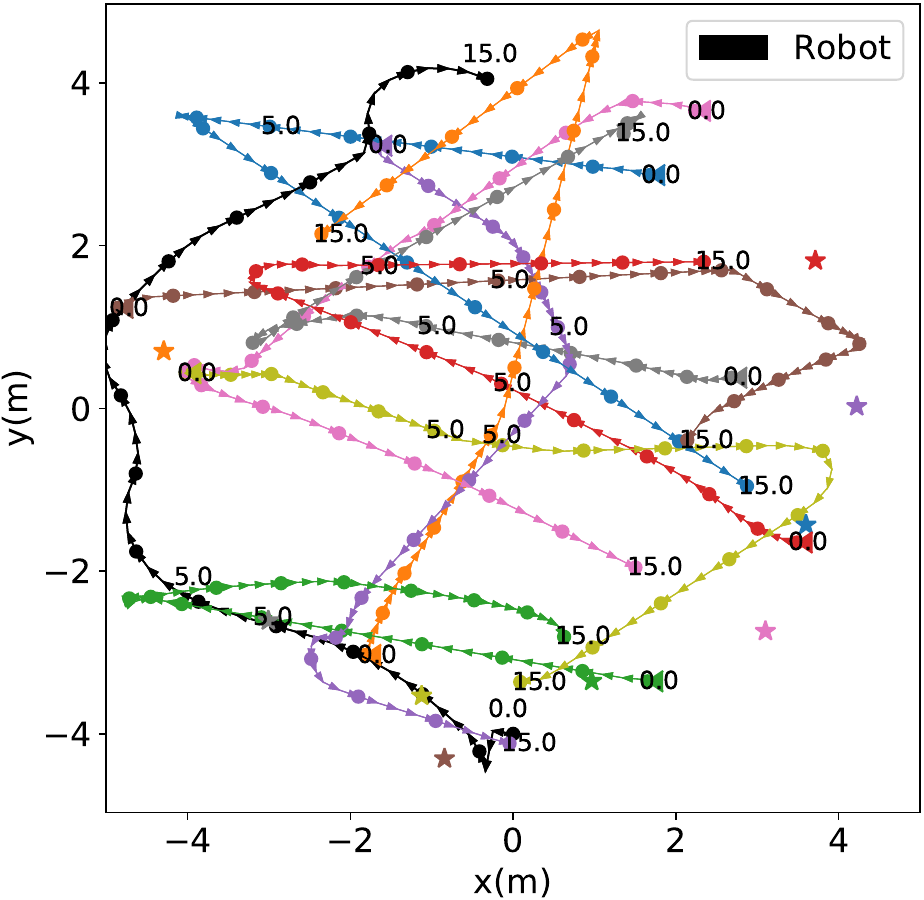}}
\subfigure[SARL]{
\label{sarlc} \includegraphics[width = 3.19cm, height = 3.19cm]{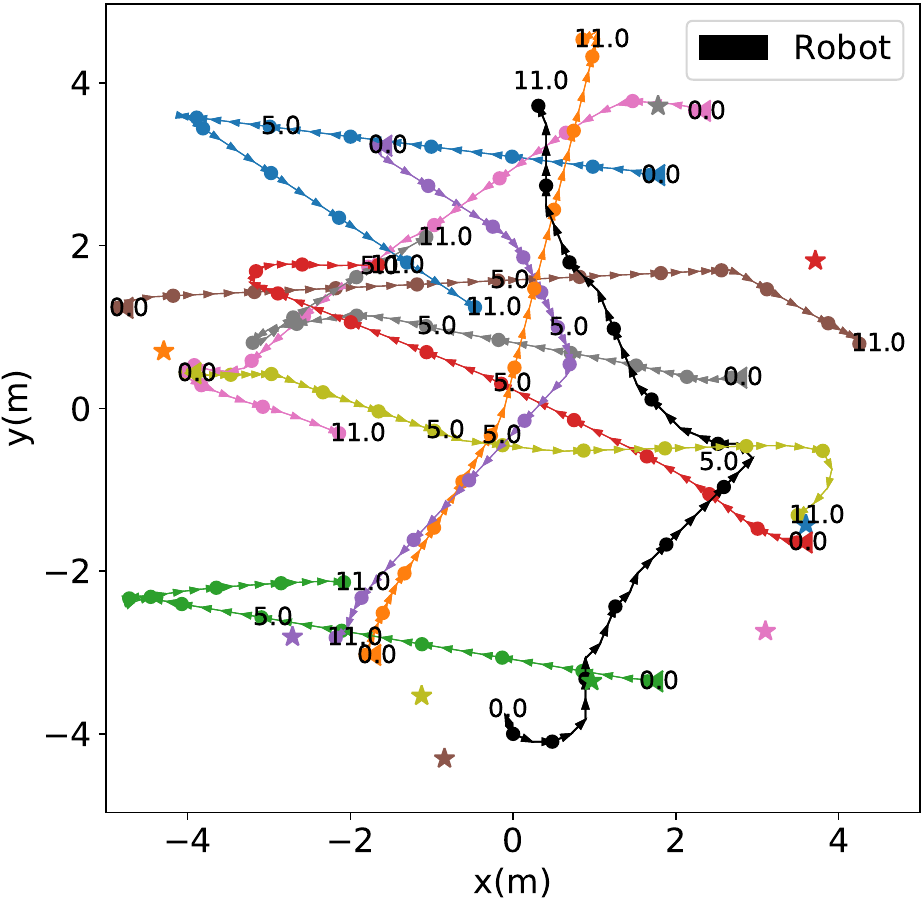}}
\subfigure[ST2-ORL]{
\label{st2c} \includegraphics[width = 3.19cm, height = 3.19cm]{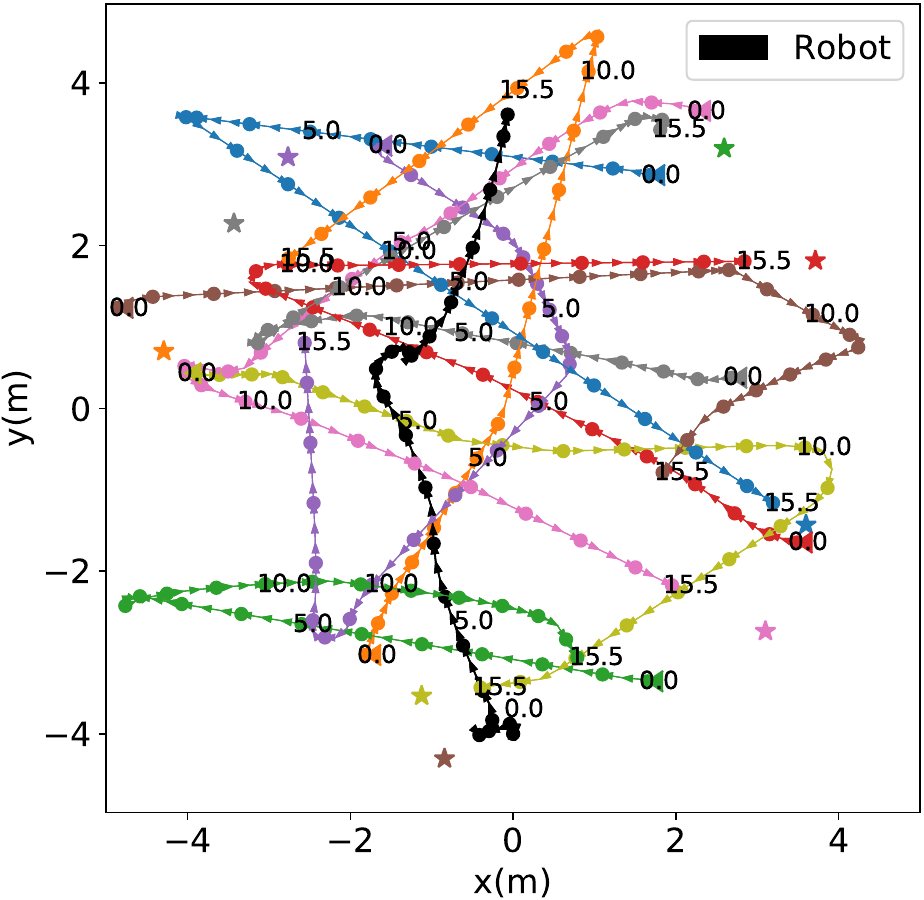}}
\caption{Illustration of trajectory in complex scenarios}
\label{complexp}
\end{figure*}

\subsection{Qualitative Evaluation}
This paper also conducts a qualitative evaluation with all baselines to further validate the effectiveness of the proposed algorithm in dynamic crowd scenarios. The trajectories of all methods are shown in Fig. \ref{simplep} and \ref{complexp}. The navigation trajectories of various methods are compared in the same setting where pedestrian policies are identical. 

Comparing figures (a)-(c) and (h) in Fig. \ref{simplep} and \ref{complexp}, it is evident that the motion trajectories generated by our algorithm are more efficient and smoother. The conservative behavior of ORCA leads to longer navigation time due to excessive avoidance of pedestrians. The limitations of the BC confine it to learning an approximate strategy similar to ORCA, which is also conservative. IQL often struggles to correctly interpret interactive information in the environment. For instance, as shown in Fig. \ref{iqls} and Fig. \ref{st2s}, IQL encounters the blue pedestrian after approximately 5 seconds and immediately moves in the opposite direction to avoid the pedestrian. In contrast, our algorithm bypasses the pedestrian and moves directly toward the goal. These trajectories also provide evidence that our algorithm, combining implicit Q-learning and spatial-temporal state transformers, is capable of capturing and comprehending spatial-temporal features utilizing the offline pedestrian-robot interactions in dynamic environments and making efficient decisions. 

As demonstrated by the last four trajectories in Fig. \ref{simplep} and \ref{complexp}, our algorithm exhibits superior trajectory behavior even when relying solely on suboptimal historical experiences. These trajectories have been proven to be better than CADRL and LSTM-RL. While DS-RNN demonstrates excellent path planning in simple scenarios, in dealing with more complex crowds, as shown in Fig. \ref{rnnc}, the robot tends to avoid crowded areas, resulting in less optimal paths. Although SARL achieves more favorable navigation times in both scenarios, our algorithm plans smoother navigation trajectories. These results collectively indicate that our proposed offline reinforcement learning based robot crowd navigation algorithm can effectively recover excellent strategies from suboptimal crowd navigation experiences by modifying the SARSA-style objective, equivalent to or even surpassing online algorithms.

\subsection{Deployment on Simulation Platform}
To validate the practicality of our proposed algorithm, we also simulated a real-world crowd environment on the Gazebo platform to test the navigation capabilities of the robot. In this simulated environment, the positions of pedestrians are randomly generated, and their movements are controlled by ORCA. The mobile robot, TurtleBot3 Waffle, is deployed with our algorithm to execute navigation tasks. A visual video of the experimental process can be viewed at https://youtu.be/kJVCYFhA6Z8.

\section{CONCLUSION}\label{5}
This paper introduces an offline reinforcement learning based robot crowd navigation. The proposed algorithm eliminates the need to query robot behaviors that are beyond the distribution range of pre-collected crowd navigation experiences by utilizing a SARSA-style objective loss function. It combines implicit Q-learning and spatial-temporal state transformers, enabling robots to effectively learn and capture spatial-temporal features from offline pedestrian-robot interactions. Experimental results demonstrate that our algorithm outperforms state-of-the-art offline baselines in both quantitative and qualitative evaluations. And compared to state-of-the-art DRL baselines, our algorithm achieves excellent performance in evaluation metrics and holds a distinct advantage in mitigating exploration risks and improving sampling efficiency.

\bibliographystyle{unsrtnat}  
\bibliography{Manuscript.bib}

\end{document}